\documentclass{article}
\PassOptionsToPackage{numbers, compress}{natbib}
\usepackage[preprint]{neurips_2025}

\usepackage[T1]{fontenc}
\usepackage{xcolor}
\usepackage{graphicx}
\usepackage{xspace}
\usepackage{url}

\definecolor{linkColor}{rgb}{0.2,0.4,0.6}
\definecolor{abstractbg}{gray}{0.95}
\definecolor{darkgreen}{rgb}{0.0,0.5,0.0}

\usepackage[colorlinks=true,
            linkcolor=linkColor,
            citecolor=linkColor,
            filecolor=linkColor,
            urlcolor=linkColor]{hyperref}

\usepackage{multirow}
\usepackage{booktabs}
\usepackage{amsmath,amsfonts}
\usepackage{nicefrac}
\usepackage{microtype}
\usepackage{caption}
\usepackage{subcaption}
\usepackage{makecell}
\usepackage{array}
\usepackage{float}
\usepackage{tabularx}
\usepackage{ragged2e}
\usepackage{tcolorbox}

\usepackage[utf8]{inputenc} 
\usepackage{arabtex}
\setcode{utf8}

\definecolor{linkColor}{rgb}{0.2,0.4,0.6}
\definecolor{abstractbg}{gray}{0.95}
\definecolor{darkgreen}{rgb}{0.0,0.5,0.0}

\newtcolorbox{markdownBox}{
    colback=gray!20,
    colframe=black,
    breakable,
    arc=3mm,
    title={},
    toptitle=0mm,
    bottomtitle=0mm,
    colbacktitle=gray!20,
    coltitle=black,
    fonttitle=\bfseries,
}

\makeatletter
\renewenvironment{abstract}{%
  \par\vskip 0.1in
  \begin{tcolorbox}[
    colback=abstractbg, colframe=abstractbg,
    arc=3mm, boxrule=0pt,
    left=6mm, right=6mm, top=4mm, bottom=4mm
  ]%
}{%
  \end{tcolorbox}%
}
\makeatother

\def\eqref#1{equation~\ref{#1}}

\def\1{\bm{1}}

\DeclareMathAlphabet{\mathsfit}{\encodingdefault}{\sfdefault}{m}{sl}
\SetMathAlphabet{\mathsfit}{bold}{\encodingdefault}{\sfdefault}{bx}{n}

\newcommand\our{\textsc{Hala}}
\newcommand\ourseries{\textsc{Hala}}
\newcommand\halas[1]{\textsc{Hala-#1}}

\newcommand{\huggingface}{\raisebox{-1.5pt}{\includegraphics[height=1.05em]{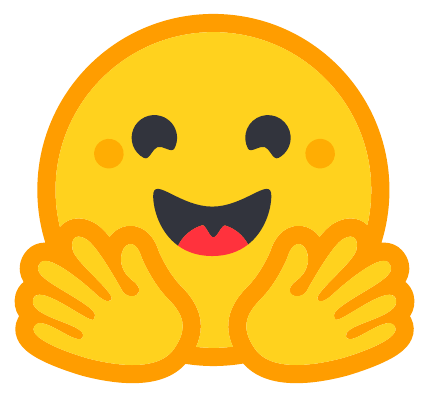}}\xspace}
\newcommand{\github}{\raisebox{-1.5pt}{\includegraphics[height=1.05em]{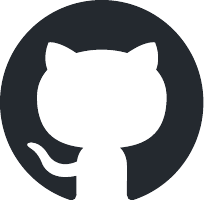}}\xspace}

\newcommand*\samethanks[1][\value{footnote}]{\footnotemark[#1]}

\title{\our\ Technical Report: Building Arabic-Centric Instruction \& Translation Models at Scale}

\author{
Hasan Abed Al Kader Hammoud\thanks{Equal Contribution. Correspondence to \texttt{hasanabedalkader.hammoud@kaust.edu.sa}} \qquad
Mohammad Zbeeb\samethanks \qquad
Bernard Ghanem\\
King Abdullah University of Science and Technology
}

\begin{document}

\maketitle

\begin{figure}[h!]
    \centering    \includegraphics[width=0.6\linewidth]{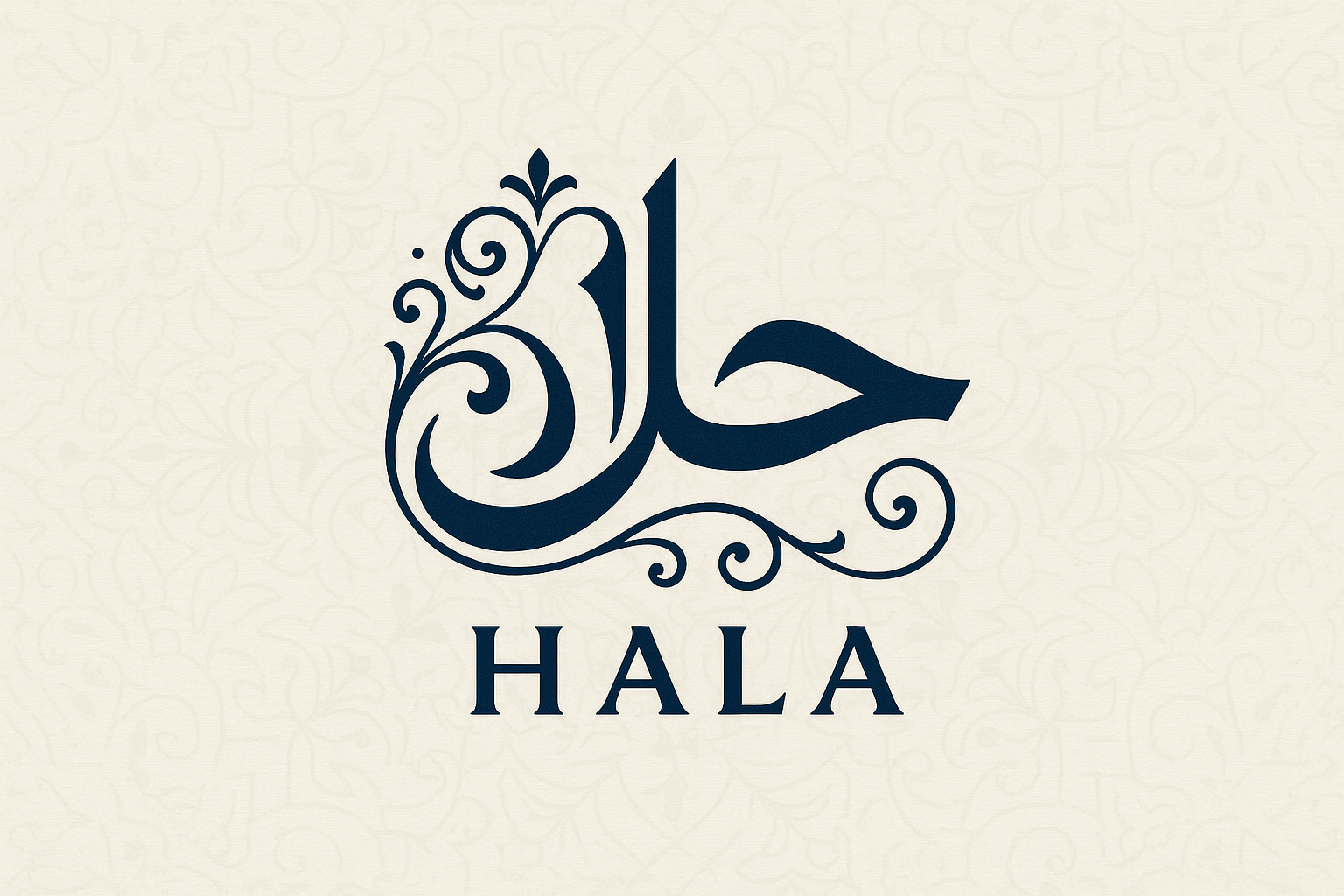}
    \label{fig:placeholder}
\end{figure}

\begin{quote}\small\itshape
\textcolor{linkColor}{\textbf{In Arabic, \RL{حلا}~(Hala) conveys sweetness and beauty - qualities long associated with the language itself.  
In this spirit, we call our models \textit{Hala}.}}
\end{quote}

\begin{abstract}
We present \our{}, a family of \emph{Arabic-centric} instruction and translation models built with our translate-and-tune pipeline. We first compress a strong AR$\leftrightarrow$EN teacher to FP8 (yielding $\sim$2$\times$ higher throughput with no quality loss) and use it to create high-fidelity bilingual supervision. A lightweight language model \texttt{LFM2-1.2B} is then fine-tuned on this data and used to translate high-quality English instruction sets into Arabic, producing a million-scale corpus tailored to instruction following. We train \our{} models at 350M, 700M, 1.2B, and 9B parameters, and apply \texttt{slerp} merging to balance Arabic specialization with base-model strengths. On Arabic-centric benchmarks, \our{} achieves state-of-the-art results within both the “nano’’ ($\leq$2B) and “small’’ (7--9B) categories, outperforming their bases. We release models, data, evaluation, and recipes to accelerate research in Arabic NLP.
\end{abstract}

\vspace{-0.5em}
\begin{table}[H]
\centering
\begin{tabular}{@{}r@{\hspace{2pt}}l@{}}
\huggingface & \textbf{Models \& Data}: \href{https://huggingface.co/collections/hammh0a/hala-68bf02b34a14b9f22305ab3a}{\texttt{hf.co/collections/Hala}} \\
\github & \textbf{Code}: \href{https://github.com/hammoudhasan/Hala}{\texttt{github.com/Hala}} \\
\end{tabular}
\end{table}

\begin{figure}[!ht]
\centering
\includegraphics[width=\linewidth]{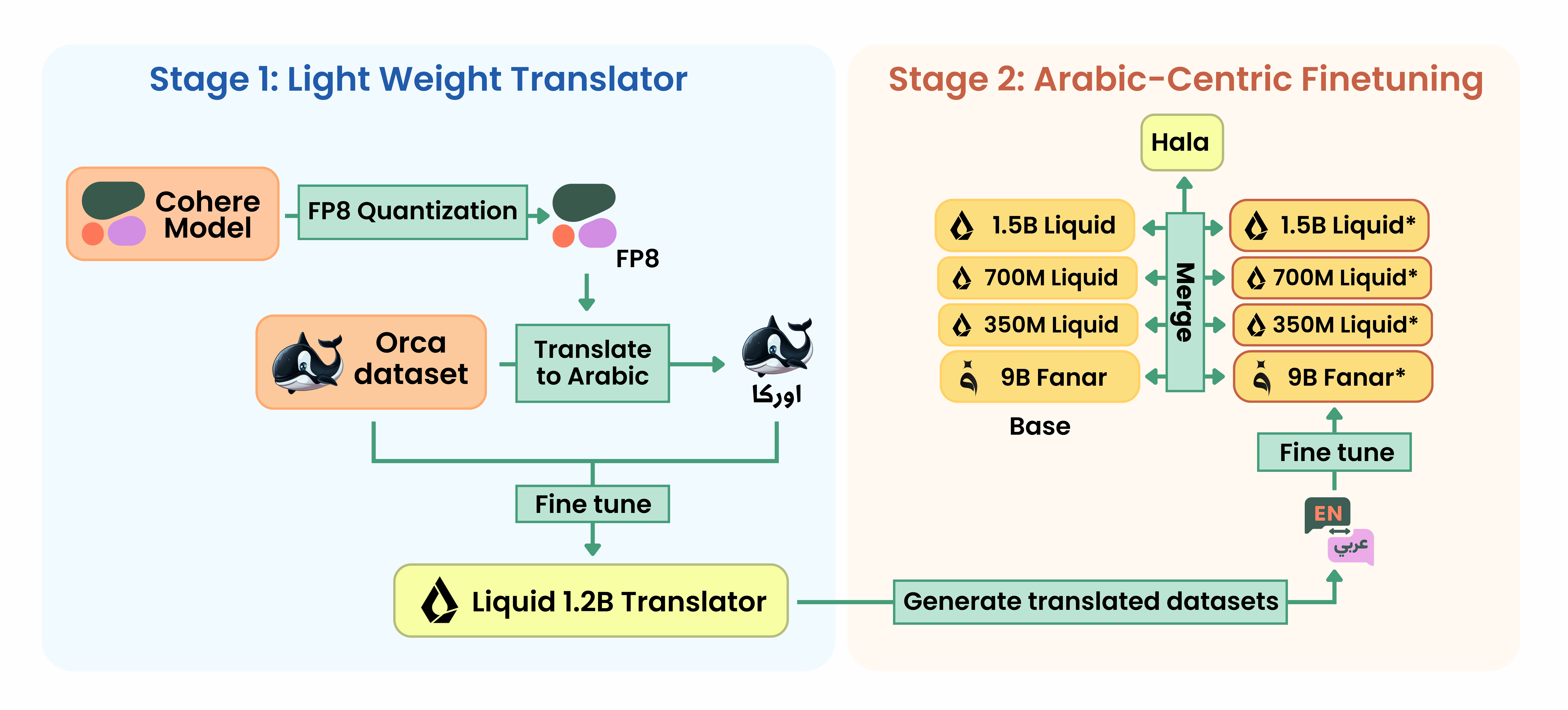}

\caption{Cross-lingual translation and fine-tuning pipeline for Liquid 1.2B. In the teacher phase, the Cohere model with FP8 inference is used to translate the Orca dataset, which is then used to fine-tune Liquid 1.2B. In the bootstrapped translator phase, Liquid 1.2B translates datasets, producing group of arabic dataset. Liquid models and FANAR were then further fine-tuned on the combined translated  datasets, yielding the final instruction-tuned models. (\textit{ \RL{ اوركا }  is orca in arabic})}
\label{fig:hala-pipeline}
\end{figure}

\section{Introduction}

Large language models (LLMs) have rapidly advanced the state-of-the-art across general-purpose NLP, demonstrating strong capabilities in few-shot learning, instruction following, and multistep reasoning. Early milestones such as GPT-3 \cite{brown2020languagemodelsfewshotlearners} catalyzed this progress, while more recent families (e.g., Gemini \cite{team2023gemini}, Claude~3) continue to expand the frontier of capability and reliability. Open-weight counterparts, including DeepSeek \cite{liu2024deepseek}, LLaMA~3 \cite{grattafiori2024llama3herdmodels}, Qwen \cite{yang2025qwen3technicalreport}, Gemma \cite{gemmateam2025gemma3technicalreport}, and Kimi~K2 \cite{team2025kimi}, have enabled broad experimentation and downstream applications, accelerating community research into scaling, alignment, and efficient deployment.

\paragraph{Multilingual modeling at scale.}
Alongside raw capability, a major thrust in recent work targets \emph{multilinguality}: building models and resources that operate across many languages. Dataset efforts range from broad-coverage sentence-aligned corpora such as Tatoeba \cite{tiedemann2020tatoebatranslationchallenge} to large-scale conversational resources such as MASSIVE \cite{fitzgerald2022massive1mexamplemultilingualnatural}. Engineering pipelines (e.g., warc2text extraction and parallel translation) have been used to derive multilingual corpora from web archives \cite{degibert2024newmassivemultilingualdataset}. Beyond data, analyses probe whether models preserve knowledge and answer consistency across languages \cite{ifergan2024beneathsurfaceconsistencyexploring}. Model design has also embraced multilinguality from the ground up: BLOOM \cite{workshop2022bloom} supports 46 languages, while Baichuan-2 \cite{yang2023baichuan} and other families emphasize improved performance on non-English tasks. Despite this breadth-first progress, per-language depth and cultural alignment remain uneven, especially for underrepresented languages.

\paragraph{Arabic LLMs and the instruction-data bottleneck.}
Arabic poses distinct challenges due to diglossia, rich morphology, and wide dialectal variation. A growing line of Arabic-centric work \cite{alkhalifa2025landscapearabiclargelanguage} spans monolingual pre-training (e.g. AraBERT \cite{antoun2021araberttransformerbasedmodelarabic}), foundation and chat models (e.g. JAIS \cite{sengupta2023jaisjaischatarabiccentricfoundation}, FANAR \cite{fanarteam2025fanararabiccentricmultimodalgenerative}, PEACOCK \cite{alwajih2024peacockfamilyarabicmultimodal}, ACE-GPT \cite{huang2023acegpt}, ALLAM \cite{bari2024allam}) and broader sovereign AI efforts such as Falcon \cite{almazrouei2023falconseriesopenlanguage}. Benchmarks, including Arabic-MMLU \cite{koto2024arabicmmlu}
, provide initial evaluation scaffolding, although coverage and difficulty remain limited relative to English. A persistent bottleneck is the scarcity of \emph{high-quality Arabic instruction data}, which constrains both instruction tuning and scaling. Previous works document the underrepresentation of non-English languages in pretraining corpora and their impact on downstream performance \cite{lin2022fewshotlearningmultilinguallanguage,xue-etal-2021-mt5,touvron2023llama2openfoundation}. In parallel, the community has explored the paradigms of 'AI trains AI', e.g., self-instruction and synthetic supervision, to overcome data scarcity \cite{xu2023wizardlm,mukherjee2023orca,achiam2023gpt,wang2022self}. However, in Arabic, the volume and fidelity of the instruction data still lag behind.

\paragraph{Language-centric vs.\ multilingual.}
We adopt the term \emph{language-centric} to denote models whose \emph{primary optimization target} is depth of capability in a specific language (here, Arabic), rather than uniform breadth across many languages. A language-centric approach can better capture linguistic nuance (e.g. morphology, orthography)\cite{conneau-etal-2020-unsupervised}, dialectal variation, and cultural/safety alignment, while still benefiting from cross-lingual transfer when appropriate. In practice, this requires (i) reliable translation pipelines to convert strong English supervision into Arabic \emph{without} eroding instruction fidelity, and (ii) training strategies that scale across model sizes while preserving Arabic fluency and task competence.

\paragraph{LLMs as translators: opportunities for Arabic data bootstrapping.}
LLMs have recently emerged as strong machine translation engines \cite{lyu2024paradigm}, capable of long-document and stylistic translation, interactive workflows, and even domain-preserving scientific translation \cite{kleidermacher2025science}. Creative strategies, such as searching for keywords / topics with multiple generations of candidates and selection \cite{he2023exploringhumanliketranslationstrategy}, further improve quality. Broad evaluations in 120+ languages \cite{zhu2024multilingualmachinetranslationlarge} suggest that carefully managed LLMs can serve as reliable translators. These developments make \emph{translation-first} bootstrapping especially attractive for Arabic instruction tuning: If we can (1) compress a capable translator for efficient, scalable inference and (2) preserve instruction semantics during translation, we can unlock large Arabic corpora suitable for high-quality tuning.

\paragraph{Our approach and contributions.}
In this report, we introduce \our{}, a family of Arabic \emph{language-centric} instruction and translation models built around an efficient translate-and-tune pipeline. Our contributions are as follows:

\begin{itemize}
    \item \textbf{Lightweight AR{\(\leftrightarrow\)}EN translator.} We compress a strong multilingual translator to FP8 with dynamic scaling using \textit{LLM Compressor} \cite{llmcompressor2024} and fine-tune \texttt{LiquidAI/LFM2-1.2B} to serve as a fast, robust AR{\(\leftrightarrow\)}EN engine. This translator is used to construct Arabic instruction data at scale while maintaining fidelity to the source instructions.
    \item \textbf{Million-scale bilingual supervision.} We build a 1.25M AR{\(\leftrightarrow\)}EN bilingual corpus by pairing translated and original texts (e.g., from Open-Orca \cite{mukherjee2023orca}) and a filtered subset of OPUS-100 \cite{zhang-etal-2020-improving}, enabling stable training of lightweight translation models and consistency checks.
    \item \textbf{Large Arabic instruction corpus.} Using our translation stack, we convert several high-quality English instruction datasets into Arabic, including Hermes~3 \cite{teknium2024hermes3technicalreport}, SCP-116K \cite{lu2025scp}, ReAlign-Alpaca \cite{fan2024reformatted}, LaMini \cite{wu-etal-2024-lamini}, Tulu~3 \cite{lambert2024tulu3}, and Synthetic Instruct-GPT-J Pairwise \cite{alex_havrilla_2023}, alongside Open-Orca \cite{mukherjee2023orca}. The resulting Arabic corpus (millions of pairs) emphasizes instruction following, reasoning, and alignment.
    \item \textbf{Arabic-centric models across scales.} We release \our{} models at 350M, 700M, and 1.2B parameters (based on Liquid checkpoints) as well as a 9B model built on the FANAR architecture \cite{fanarteam2025fanararabiccentricmultimodalgenerative}. To combine complementary strengths from English- and Arabic-tuned checkpoints, we employ \textit{MergeKit} \cite{mergekit} with spherical linear interpolation.
    \item \textbf{Open releases and recipes.} We release models, data, and training/evaluation scripts to facilitate reproducibility and further research on Arabic instruction tuning.
\end{itemize}

\paragraph{Summary.}
By coupling an efficient AR{\(\leftrightarrow\)}EN translator with million-scale data construction, \our{} advances Arabic instruction tuning under constrained compute budgets. Our results (Section~\ref{sec:evaluation}) indicate that \our{} models achieve competitive performance within their parameter classes on Arabic-centric benchmarks \cite{koto2024arabicmmlu}, supporting the view that \emph{language-centric} modeling is a practical and effective complement to breadth-first multilingual scaling.

\section{Methodology}
\label{sec:methodology}

\subsection{Quantizing the main translator to FP8}
We begin with a high-capacity multilingual translator (\texttt{CohereLabs/command-a-translate-08-2025}) and compress it to FP8 \cite{kuzmin2022fp8} with \emph{dynamic scaling} using \textit{LLM Compressor} \cite{llmcompressor2024}, releasing the FP8 artifact as \texttt{hammh0a/command-a-translate-FP8-Dynamic}. The FP8 conversion reduces memory footprint and improves inference throughput (empirically \(\approx\!2\times\) faster than the non-quantized counterpart) while preserving translation quality on our evaluation sets. We follow the official \texttt{llm-compressor} recipe (per-tensor dynamic scaling and post-conversion validation) to ensure stability.

\subsection{Bootstrapping bilingual supervision from Open-Orca}
To construct high-quality AR{\(\leftrightarrow\)}EN supervision aligned with instruction-tuning style data, we translate the \emph{first 405K} instruction--response pairs from \texttt{Open-Orca/OpenOrca} \cite{mukherjee2023orca} \emph{into Arabic}, covering both the user questions and assistant responses. The quantized FP8 translator is prompted with a minimal instruction:

\begin{quote}
\small
\ttfamily
Translate from English to Arabic: \{x\}
\end{quote}

For each example, we \emph{pair} the Arabic translations with their original English counterparts, yielding bilingual tuples of the form
\(\langle \text{instr}_{\text{en}}, \text{instr}_{\text{ar}}, \text{resp}_{\text{en}}, \text{resp}_{\text{ar}} \rangle\).
This produces an instruction-focused bilingual set mirroring the semantics and difficulty of \texttt{Open-Orca}, with substantial coverage of reasoning-heavy queries.

\subsection{Quality filtering of OPUS-100 with a strict bilingual judge}
We augment the above with a large parallel corpus drawn from the \texttt{Helsinki-NLP/opus-100} \cite{zhang-etal-2020-improving} \texttt{ar-en} subset. From 1M candidate pairs, we filter for fidelity using a compact judge model (\texttt{Qwen2.5-3B-Instruct} \cite{yang2025qwen3technicalreport}) prompted to emit a binary verdict (\texttt{accept}/\texttt{reject}):

\begin{verbatim}
prompt = f"""
You are a strict bilingual judge. You will be given a translation pair.
Arabic: {ar_text}
English: {en_text}

If the English is a correct and natural translation of the Arabic, output only:
accept
Otherwise, output only:
reject
"""
\end{verbatim}

Pairs marked \texttt{accept} are retained; this procedure yields \textbf{439{,}592} accepted pairs out of \(\sim\)1M candidates, providing a clean AR{\(\leftrightarrow\)}EN signal complementary to the instruction-style data above.

\subsection{Training a lightweight AR{\(\leftrightarrow\)}EN translator}
We combine the translated \texttt{Open-Orca} set (405K $
\times$ 2 = 810K) with the filtered \texttt{OPUS-100} pairs (440K), totaling \(\sim\)1.26M bilingual examples, and fine-tune \texttt{LiquidAI/LFM2-1.2B} into a fast, stable AR{\(\rightarrow\)}EN translator specialized for instruction-style inputs (instructions and responses). We use simple chat-style prompting during training (for E{\(\rightarrow\)}A) and standard supervised fine-tuning with cross-entropy. This lightweight translator serves as \emph{workhorse} for the construction of large-scale Arabic data in the next stage.

\subsection{Building the Arabic instruction corpus via translation}
Using the above translator, we convert multiple high-quality English instruction datasets into Arabic, preserving formatting and answer style:
\begin{itemize}
    \item \textbf{Open-Orca/OpenOrca} \cite{mukherjee2023orca}: 405K (first subset), covering multi-step, reasoning-heavy queries.
    \item \textbf{NousResearch/Hermes-3-Dataset} \cite{teknium2024hermes3technicalreport}: filtered to remove all code-related samples to avoid translation artifacts.
    \item \textbf{EricLu/SCP-116K} \cite{lu2025scp}: instructional and conversational pairs.
    \item \textbf{GAIR/ReAlign-Alpaca} \cite{fan2024reformatted}: realigned version of Alpaca instructions.
    \item \textbf{Dahoas/synthetic-instruct-gptj-pairwise} \cite{alex_havrilla_2023}: synthetic paired preference-style instructions.
    \item \textbf{MBZUAI/LaMini-instruction} \cite{wu-etal-2024-lamini}: lightweight instruction data, translated fully.
    \item \textbf{allenai/tulu-3-sft-mixture} \cite{lambert2024tulu3}: we keep only English subsets and translate them.
\end{itemize}
The resulting corpus emphasizes instruction following, reasoning, and alignment, providing broad coverage for Arabic-centric instruction tuning. We collect a total of roughly $4.5$M samples.

\subsection{Arabic instruction fine-tunes and model merging}
We fine-tune models across scales on the translated Arabic instruction mix, then apply merging to balance Arabic gains with base-model strengths:
\begin{itemize}
    \item \textbf{350M}: fine-tune \texttt{LiquidAI/LFM2-350M}, then merge with its base to obtain \halas{350M}.
    \item \textbf{700M}: fine-tune \texttt{LiquidAI/LFM2-700M}, then merge with its base to obtain \halas{700M}.
    \item \textbf{1.2B}: fine-tune \texttt{LiquidAI/LFM2-1.2B}, then merge with its base to obtain \halas{1.2B}.
    \item \textbf{9B}: fine-tune on top of \texttt{QCRI/Fanar-1-9B-Instruct} \cite{fanarteam2025fanararabiccentricmultimodalgenerative}, then merge with its base to obtain \halas{9B}.
\end{itemize}
Merging is performed with \textit{MergeKit} \cite{mergekit} using spherical linear interpolation (\texttt{slerp}) at \(t{=}0.5\), which we found to preserve general capability while boosting Arabic instruction-following performance. The overall translate--and--tune pipeline is illustrated in Fig.~\ref{fig:hala-pipeline}.

\section{Evaluation}
\label{sec:evaluation}

\paragraph{Benchmarks and protocol.}
We evaluate on a suite of Arabic-centric tasks following the \emph{Open-Arabic-LLM-Leaderboard (OALL)} task selection where feasible. Concretely, we report results on:
\textbf{AlGhafa} \cite{almazrouei-etal-2023-alghafa}, \textbf{AraTrust} \cite{aratrust}, \textbf{ArabicMMLU} \cite{koto2024arabicmmlu},
\textbf{ArbMMLU-HT} \cite{koto2024arabicmmlu}, \textbf{EXAMS} \cite{hardalov-etal-2020-exams}, and \textbf{MadinahQA} \cite{koto2024arabicmmlu}.
We \emph{exclude} \textbf{Alrage} (present in some OALL variants) because it requires an LLM-as-a-judge setup.
All evaluations are conducted with \texttt{LightEval} \cite{lighteval} using \texttt{vLLM} \cite{kwon2023efficient} as the backend for efficient, reproducible inference. We will release exact \emph{LightEval} command lines, task definitions, and configuration files in the accompanying GitHub repository (\github).

\paragraph{Model families.}
To contextualize \our{} within the broader landscape, we include models spanning both multilingual and Arabic-centric families:
LLaMA \cite{grattafiori2024llama3herdmodels},
Qwen \cite{yang2025qwen3technicalreport},
Gemma \cite{gemmateam2025gemma3technicalreport},
SILMA \cite{silma-9b-2024, silma-kashif-2b-2024},
Saka,
FANAR \cite{fanarteam2025fanararabiccentricmultimodalgenerative},
Yehia \cite{yehia2025},
ALLAM,
Command-R, and LiquidAI.
We report our \ourseries{} models at 350M, 700M, 1.2B, and 9B parameters alongside their corresponding bases (LiquidAI checkpoints and FANAR), and representative competitive baselines (e.g., Command-R-7B Arabic).

\paragraph{Main results.}
The aggregated results across the six benchmarks are summarized in Table~\ref{tab:main-results}. In the \emph{nano} regime (\(\leq\)2B), \halas{1.2B} improves substantially over its base (LiquidAI/LFM2-1.2B), achieving the best average within the size bucket (cf.\ Table~\ref{tab:main-results}). Similarly, \halas{350M} and \halas{700M} consistently outperform their Liquid bases across most tasks, indicating that our translate--and--tune pipeline yields \emph{consistent Arabic gains} even at very small scales. In the \emph{small} regime (\(\leq\)9B), \halas{9B} consistently outperforms the previous state-of-the-art \texttt{QCRI/Fanar-1-9B-Instruct} baseline on the average metric, while maintaining competitive scores on individual tasks. These trends support our central claim: \emph{language-centric} tuning on high-fidelity Arabic instruction data improves Arabic capability across scales, and merging \cite{sanyal2023earlyweightaveragingmeets} (\texttt{slerp}, \(t{=}0.5\)) preserves general competence while enhancing Arabic instruction-following.
\begin{table}[t]
\centering
\renewcommand{\arraystretch}{1.2}
\begingroup
\setlength{\tabcolsep}{5.5pt}
\newcommand{\best}[1]{\textbf{#1}}
\newcommand{\secondbest}[1]{\underline{#1}}
\newcommand{\NA}{\textemdash}

\caption{Arabic-centric evaluation across six benchmarks following the OALL task suite (excluding \emph{Alrage}); higher is better. 
Columns 4--9 report task scores (\%). \textbf{Average} is the unweighted mean across the six tasks. 
Best \textbf{Average} within each size bucket is \best{bold}; second-best is \secondbest{underlined}. 
All runs use \texttt{LightEval} with \texttt{vLLM}; exact commands are released in the repo.}
\label{tab:main-results}
\small
\resizebox{\textwidth}{!}{%
\begin{tabular}{@{}l l c rrrrrrr r@{}}
\toprule
\multirow{2}{*}{\textbf{Size}} & \multirow{2}{*}{\textbf{Model Name}} & \multirow{2}{*}{\textbf{Params}} 
& \multicolumn{6}{c}{\textbf{Arabic-centric Benchmarks (\%)}} & \multirow{2}{*}{\textbf{Average}} \\
\cmidrule(lr){4-9}
& & & AlGhafa & ArabicMMLU & EXAMS & MadinahQA & AraTrust & ArbMMLU-HT & \\
\midrule
\multicolumn{10}{@{}l}{\textit{Nano (\(\leq\)2B)}} \\
\midrule
\(\leq\)2B & meta-llama/Llama-3.2-1B & 1B & 33.9 & 26.5 & 21.2 & 25.7 & 37.1 & 23.9 & 28.0 \\
\(\leq\)2B & Qwen/Qwen2-1.5B-Instruct & 1.5B & 53.1 & 49.2 & 35.2 & 45.5 & 68.9 & 37.4 & 48.2 \\
\(\leq\)2B & Qwen/Qwen2.5-1.5B-Instruct & 1.5B & 48.4 & 43.5 & 31.8 & 38.2 & 70.8 & 35.9 & 44.8 \\
\(\leq\)2B & Sakalti/Saka-1.5B & 1.5B & 51.4 & 40.0 & 31.3 & 31.5 & 47.5 & 33.5 & 39.2 \\
\(\leq\)2B & Qwen/Qwen3-1.7B-Base & 1.7B & 56.8 & 49.7 & 38.2 & 40.0 & 75.6 & 43.9 & \secondbest{50.7} \\
\(\leq\)2B & Qwen/Qwen1.5-1.8B & 1.8B & 32.7 & 26.7 & 23.8 & 26.0 & 31.5 & 23.6 & 27.4 \\
\(\leq\)2B & silma-ai/SILMA-Kashif-2B-Instruct-v1.0 & 2B & 59.7 & 45.6 & 33.1 & 38.8 & 73.3 & 35.8 & 47.7 \\
\(\leq\)2B & google/gemma-2-2b-it & 2B & 34.1 & 30.1 & 23.6 & 20.1 & 31.2 & 23.4 & 27.1 \\
\cmidrule(lr){1-10}
\(\leq\)2B & LiquidAI/LFM2-350M & 350M & 39.0 & 35.2 & 30.9 & 28.3 & 43.3 & 29.1 & 34.3 \\
\(\leq\)2B & \ourseries-350M & 350M & 51.4 & 41.2 & 36.9 & 34.5 & 52.1 & 35.4 & 41.9 \textcolor{darkgreen}{(+7.6)} \\
\cmidrule(lr){1-10}
\(\leq\)2B & LiquidAI/LFM2-700M & 700M & 50.1 & 38.3 & 34.3 & 32.5 & 56.3 & 37.2 & 41.4 \\
\(\leq\)2B & \ourseries-700M & 700M & 55.5 & 45.9 & 40.6 & 34.7 & 65.2 & 39.4 & 46.9 \textcolor{darkgreen}{(+5.5)} \\
\cmidrule(lr){1-10}
\(\leq\)2B & LiquidAI/LFM2-1.2B & 1.2B & 53.8 & 45.2 & 35.0 & 34.7 & 65.6 & 43.4 & 46.3 \\
\(\leq\)2B & \ourseries-1.2B & 1.2B & 59.2 & 48.6 & 43.4 & 41.6 & 71.7 & 44.2 & \best{51.4} \textcolor{darkgreen}{(+5.1)} \\
\midrule\midrule
\multicolumn{10}{@{}l}{\textit{Small (7B--9B)}} \\
\midrule
7B--9B & CohereForAI/c4ai-command-r7b-arabic-02-2025 & 7B & 74.8 & 59.3 & 65.0 & 63.8 & 80.5 & 50.1 & 65.6 \\
7B--9B & JasperV13/Yehia-7B-DPO-Reasoning-preview & 7B & 75.1 & 66.3 & 51.8 & 54.9 & 81.9 & 55.1 & 64.2 \\
7B--9B & Navid-AI/Yehia-7B-preview & 7B & 70.8 & 64.9 & 52.1 & 54.4 & 87.5 & 53.4 & 63.9 \\
7B--9B & JasperV13/Yehia-7B-Reasoning-preview & 7B & 75.2 & 66.3 & 52.7 & 55.0 & 80.8 & 55.2 & 64.2 \\
7B--9B & ALLaM-AI/ALLaM-7B-Instruct-preview & 7B & 69.5 & 64.9 & 51.6 & 54.2 & 86.9 & 52.8 & 63.3 \\
7B--9B & Qwen/Qwen2-7B-Instruct & 7B & 73.2 & 60.0 & 47.3 & 59.5 & 82.8 & 51.3 & 62.4 \\
7B--9B & Qwen/Qwen3-8B-Base & 8B & 74.8 & 65.0 & 52.5 & 52.2 & 83.4 & 61.5 & 64.9 \\
\cmidrule(lr){1-10}
7B--9B & QCRI/Fanar-1-9B-Instruct & 9B & 76.4 & 65.8 & 52.7 & 73.3 & 88.3 & 58.6 & \secondbest{69.2} \\
7B--9B & \ourseries-9B & 9B & 78.3 & 65.6 & 53.8 & 70.4 & 89.6 & 61.4 & \best{69.9} \textcolor{darkgreen}{(+0.7)} \\
\bottomrule
\end{tabular}}
\endgroup
\end{table}

\paragraph{Translator quality: EN\(\rightarrow\)AR MMLU question translation.}
We assess translation fidelity in an instruction-style regime by constructing a controlled, reference-based evaluation using \texttt{cais/mmlu} (English questions) and \texttt{openai/mmmlu} (Arabic questions). We uniformly sample \(n{=}500\) English questions from \texttt{cais/mmlu} with a fixed random seed, translate each to Arabic using the system under test, and align it to its ground-truth Arabic counterpart from the \texttt{openai/mmmlu} Arabic subset (same subject and item ID). We report \emph{BLEU} (SacreBLEU, 13a tokenization), \emph{ROUGE-L} (F1, \texttt{rouge-score}), and \emph{chrF++} (SacreBLEU) between the system output and the reference Arabic question. Exact sampling seeds, preprocessing, and metric commands will be released in the accompanying repository.

\textit{Prompting setup (fairness control).} To ensure comparability across systems, we use fixed prompts:
\begin{itemize}
    \item \textbf{LiquidAI/LFM2-1.2B (specialized translator) prompt}
\begin{verbatim}
You are a professional translation engine. 
Translate English to Modern Standard Arabic.
Reply ONLY with the Arabic translation—no quotes, notes, or explanations.
Translate everything that follows into Arabic: {text}
\end{verbatim}
    \item \textbf{All other models (teacher FP16/FP8 and baselines) prompt}
\begin{verbatim}
Translate everything that follows into Arabic: {text}
\end{verbatim}
\end{itemize}
Here, \texttt{\{text\}} is replaced verbatim by the English question from \texttt{cais/mmlu}. Outputs are evaluated directly against the paired Arabic reference from \texttt{openai/mmmlu} without post-processing beyond each metric’s built-in normalization.

\begin{table}[H]
\centering
\renewcommand{\arraystretch}{1.15}
\setlength{\tabcolsep}{7pt}
\small
\newcommand{\gain}[1]{\textcolor{green!50!black}{\scriptsize\,(#1)}}
\caption{\textbf{EN\(\rightarrow\)AR translation quality on 500 sampled MMLU questions.}
References come from the Arabic subset of \texttt{openai/mmmlu}. Higher is better.
Values in \gain{\,\(\cdot\)} denote absolute deltas vs.\ the reference system within each block
(FP8 vs.\ FP16 for the teacher translator; \our{} vs.\ \texttt{LFM2-1.2B} base for the lightweight translator).
Prompts are fixed as specified above.}
\label{tab:translator-eval}
\begin{tabular}{lccc}
\toprule
\textbf{System} & \textbf{BLEU}~\(\uparrow\) & \textbf{ROUGE-L}~\(\uparrow\) & \textbf{chrF++}~\(\uparrow\) \\
\midrule
\multicolumn{4}{l}{\textit{Teacher translator}} \\
\midrule
CohereLabs/command-a-translate-08-2025 (FP16) & 53.1 & 26.0 & 68.6 \\
\textbf{hammh0a/command-a-translate-FP8-Dynamic} & 53.5 \gain{+0.3} & 26.0 \gain{+0.0} & 68.9 \gain{+0.3} \\
\midrule
\multicolumn{4}{l}{\textit{Lightweight translator (LFM2-1.2B family)}} \\
\midrule
LiquidAI/LFM2-1.2B (base) & 16.0 & 19.3 & 43.2 \\
\textbf{\our{} LFM2-1.2B Translator (ours)} & 48.2 \gain{+32.1} & 25.1 \gain{+5.9} & 64.2 \gain{+21.0} \\
\bottomrule
\end{tabular}
\end{table}

\noindent\textbf{Compute and cost.} All models were trained within a budget of \${}1{,}000 on 8$\times$H100-SXM GPUs, and dataset translation was performed on 12$\times$A100 GPUs at an additional cost of roughly \${}500.

\section{Conclusion}
We presented \our{}, a family of \emph{language-centric} Arabic models that leverage an efficient translate--and--tune pipeline: compress a capable AR{\(\leftrightarrow\)}EN translator to FP8, bootstrap million-scale Arabic instruction data from high-quality English sources, and fine-tune compact and small models with \texttt{slerp}-based merging. \our{} delivers consistent improvements over base Liquid and FANAR checkpoints, achieving state-of-the-art averages in both the \emph{nano} (\(\leq\)2B) and \emph{small} (\(\leq\)9B) categories on a diverse Arabic benchmark suite. We release models, data, and recipes to catalyze further research on Arabic instruction tuning and to encourage \emph{language-centric} approaches as a complement to breadth-first multilingual scaling.

\section{Acknowledgements}

The research reported in this publication was supported by funding from King Abdullah University of Science and Technology (KAUST) - Center of Excellence for Generative AI, under award number 5940.

\bibliographystyle{IEEEtranN}
\bibliography{main}

\end{document}